\documentclass{article}
\usepackage{kr}

\usepackage{times}
\usepackage{soul}
\usepackage{url}
\usepackage[hidelinks]{hyperref}
\usepackage[utf8]{inputenc}
\usepackage[small]{caption}
\usepackage{graphicx}
\usepackage{amsmath}
\usepackage{amsthm}
\usepackage{booktabs}
\title{Multi-Valued Partial Order Plans in Numeric Planning}

\author{%
	Helal, Hayyan$^1$\and
	Lakemeyer, Gerhard$^2$\\
	\affiliations
	RWTH Aachen University$^{1;2}$\\
	\emails
	\{helal, gerhard.lakemeyer\}@kbsg.rwth-aachen.de
}

\usepackage{amssymb}
\usepackage{enumerate}
\usepackage{listings}
\usepackage[ruled, boxruled, vlined]{algorithm2e}

\newtheorem{definition}{Definition}
\newtheorem{corollary}{Corollary}
\newtheorem{theorem}{Theorem}
\newtheorem{spec}{Specification}
\newtheorem{lemma}{Lemma}
\newtheorem{example}{Example}

\newcommand{\viogeq}{\ensuremath{\sim^{\scriptscriptstyle\geq}}}
\newcommand{\violeq}{\ensuremath{\sim^{\scriptscriptstyle\leq}}}
\newcommand{\viocirc}{\ensuremath{\sim^{\scriptscriptstyle\circ}}}

\newcommand{\peq}{\vcenter{\hbox{\scalebox{0.7}{$+\!\!=$}}}}
\newcommand{\meq}{\vcenter{\hbox{\scalebox{0.7}{$-\!\!=$}}}}

\newcommand{\upprec}{\ensuremath{\pi^{\leq}}}
\newcommand{\loprec}{\ensuremath{\pi^{\geq}}}
\newcommand{\Aa}{\ensuremath{\mathbf{A}}}
\newcommand{\Z}{\ensuremath{\mathbb{Z}}}
\newcommand{\N}{\ensuremath{\mathbb{N}}}

\newcommand{\Pa}{\ensuremath{\mathfrak{P}}}
\newcommand{\A}{\ensuremath{\mathcal{A}}}
\newcommand{\Ss}{\ensuremath{\text{\normalfont S}}}
\newcommand{\Loprec}{\ensuremath{\Pi^\geq}}
\newcommand{\Upprec}{\ensuremath{\Pi^\leq}}
\newcommand{\I}{\ensuremath{\mathfrak{I}}}
\newcommand{\lin}{\ensuremath{\text{\normalfont lin}}}

\newcommand{\forallt}{\ensuremath{\text{ for all }}}
\newcommand{\existst}{\ensuremath{\text{ there exists }}}
\newcommand{\Forallt}{\ensuremath{\text{ For all }}}

\newcommand{\ort}{\ensuremath{\text{ or }}}
\newcommand{\andt}{\ensuremath{\text{ and }}}

\newcommand{\impliest}{\ensuremath{\text{ implies }}}
\newcommand{\stt}{\ensuremath{\text{ s.t. }}}

\begin{document}
	\maketitle
	\begin{abstract}
		Many planning formalisms allow for mixing numeric with Boolean effects. However, most of these formalisms are undecidable. In this paper, we will analyze possible causes for this undecidability by studying the number of different occurrences of actions, an approach that proved useful for metric fluents before. We will start by reformulating a numeric planning problem known as restricted tasks as a search problem. We will then show how an NP-complete fragment of numeric planning can be found by using heuristics. To achieve this, we will develop the idea of multi-valued partial order plans, a least committing compact representation for (sequential and parallel) plans. Finally, we will study optimization techniques for this representation to incorporate soft preconditions.
		
	\end{abstract}
	\section{Introduction}
	In classical planning scenarios, we use propositions mapped to truth values, preconditions that are Boolean propositional formulae, and effects that are activations/deactivations of certain propositions which get conjugated. Conjugation of propositions is called \textit{idempotent}, i.e., $ p \wedge p \equiv p $ for any proposition $ p $. We will call such planning scenarios \textit{idempotent} because doing a (propositional) action twice in a row instead of once does not change the plan.
	
	However, many planning scenarios are \textit{non-idempotent}. Human diet planning, for example, is conducted by choosing appropriate food items that fulfill the nutritional requirements in the diet formulation \cite{ss:18}. An eating action increments the amount of ``calories" or ``sugar", which are relevant when planning to lose weight. Notice that doing an eating action once differs from doing it twice, which means it is a non-idempotent scenario. In such cases, we require ourselves to eat less than $ n $ calories or allow ourselves to eat $ n $ grams of sugar only after doing sport that results in burning $ m $ ``calories", independently of the number of action repetitions that lead to these contributions. On the other hand, in curriculum planning, automated planning has been a helpful formalism for mapping actions, i.e., learning contents, in terms of preconditions (precedence relationships) and causal effects to find plans, i.e., learning paths that best fit the needs of each student \cite{c:15}. The idea for this paper was developed while working on an application that helps students plan their future courses. It uses process mining to derive soft preconditions from data of old students and rule-based AI to represent these together with the university's hard preconditions as a planning problem. We define the goal ``graduation" as a required number of ``credit points" in predefined fields. All actions in this scenario (courses, projects, etc.) contribute to integer linear values like ``credit points" and ``workload". Finally, many robotics-related planning scenarios with costs and/or resources are non-idempotent.
	
	Motivated by these scenarios, we will study the differences between idempotent and non-idempotent scenarios since propositional and numeric planning approaches can both model them but with differences in compactness and efficiency. Let us go back to our first insight: The effects of successive actions in non-idempotent planning are summed, while in the idempotent case, the effects are activated/deactivated and conjugated. For example, there is an intuitive difference between the effects of the actions \textit{move\_one\_floor\_up} and \textit{open\_the\_door}. This is reflected directly by the different occurrences/copies/repetitions of an action in idempotent and non-idempotent scenarios. If a minimal plan begins with the action \textit{open\_the\_door}, there is no need to repeat this action later on in that plan, as long as none of its effects is deactivated by a following action, like \textit{close\_the\_door}. However, in non-idempotent planning, one can use an action more than once for different reasons. For instance, it is useful to have \textit{move\_one\_floor\_up} twice in a plan, without any of its effects getting ``deactivated" or ``reversed" by applying an action like \textit{move\_one\_floor\_down}. Since (linear) numeric planning can model propositional planning, two different kinds of action repetition in (linear) numeric planning arise, one that is of \textit{linear} nature and another of \textit{Boolean} nature. We will study the differences between these two kinds to find a way to represent numeric plans in a least-committing and compact way at the same time, which will help solve the planning problem.
	\begin{figure}
		\centering
		\includegraphics[scale=0.09]{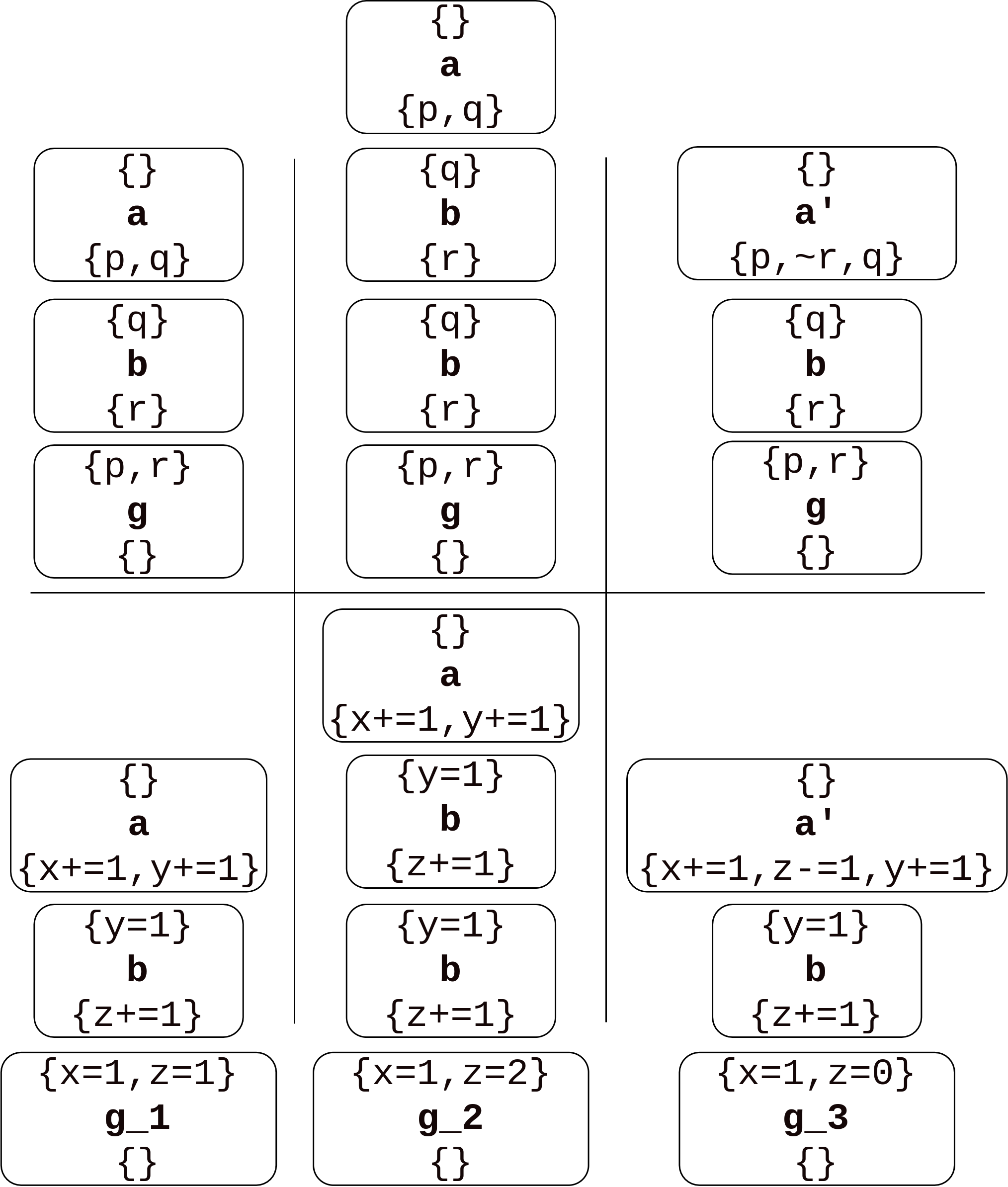}
		\caption{Demonstration of the expressive power of non-idempotent planning (below) compared to idempotent planning (above). Compare $ p, q, r $ with $ x, y, z $, respectively, in the effects and the preconditions of each action, where e.g., $ p \equiv \bot \leftrightarrow x = 0 $.}
		\label{fig:idem}
	\end{figure}
	
	In Figure 1, we show six plans (ordered top-down), three propositional (above), and three numeric (below). All actions are described by their preconditions above the action name and effects below it. We mean by $ x \peq 1 $ that the action increments $ x $ by 1 after applying it, and $ x \meq 1 $ means that the value is decremented by $ 1 $, similar to the python commands used for such operations, and the notation used by Hoffman \shortcite{h:03} for restricted tasks. Initially, $ p, q, r $ are mapped to $ \bot $, and $ x, y, z $ to $ 0 $. Each column has a certain pattern: We begin with a simple plan on the left, repeat the action $ b $ in the middle, and modify $ a $ to include a negative effect on the right. If we compare the goals satisfied by each plan, we see that all three propositional plans achieve the same goal $ g $, while three different goals are achieved by the numeric plans. These examples sum up the additional expressive power of numeric over propositional planning.
	
	\textbf{Paper outline:} We start by defining the main planning problem in Section 3, then the main solution representation in Section 4. In Section 5, we introduce an algorithm to solve the planning problem. After that, we define new heuristics in Section 6 and an NP-complete fragment of numeric planning in Section 7. Finally, we show how to optimize the resulting plans in Section 8.
	
	\section{Related Work}
	Many of the first approaches for (propositional) planning used search algorithms. However, several advancements in AI planning have been made after propositional planning and past search-based approaches. These concern the formulation as well as the solving algorithm.
	
	To name a few, PDDL2.1 \cite{fl:03}, restricted tasks \cite{h:03}, which is equivalent to Structurally Restricted Numeric Planning (SNP) \cite{shs:16}, and their super-set \cite{sgj:22}, as well as generalized planning \cite{jsj:19}, and Generalised Linear Integer Numeric Planning (GLINP) \cite{lcfgls:22} can all model non-idempotent planning scenarios described above, but the introduction of numerical variables in these formulations makes the problem undecidable \cite{h:02}. Several decidable subsets have been defined. For example, Shleyfman et al. \shortcite{sgj:22} show how SNP with restrictions on the causal structure can be PSPACE, similar to our restrictions that make the problem NP. Additionally, Qualitative Numeric Planning (QNP) \cite{szig:11} is an EXPTime-complete \cite{bg:20} planning formulation that allows for cycles and output algorithm-like plans. Plans, in our case, are inefficient for modeling cycles, as we will discuss later. However, the formulation defined in this paper sheds new light on the decidable fragments of numeric planning. In the future work section, we describe how the formulation can be extended to allow for cycles.
	
	On the advancements of solving algorithms, Integer Linear Programming (ILP) has shown helpful for AI planning because it allows for the incorporation of numeric conditions to AI planning \cite{vbln:99}. We will show the usefulness of this for plan optimization. However, most ILP approaches output linear plans, e.g., \cite{netal:05}, which is more challenging and too committing. We only need certain causal relations for the plan to hold \cite{h:04}. We will use ILP to calculate a multi-valued partial order plan representing these causal relations and show how that reduces the search space. Additionally, it will allow for the easy incorporation of soft preconditions within planning.
	
	However, not only ILP but also heuristic search has shown very useful for numeric planning, for instance, by using abstraction methods, \cite[ and others]{sh:14,ls:17}. Hoffmann \shortcite{h:03} proposed an extension of delete lists, a relaxation of propositional planning, to numeric planning. However, that paper deals with ``acyclic $ := $-effects" (e.g., the effect $ x := 2 $, which sets the value of $ x $ to $ 2 $), while we focus on a broader term, ``acyclic violation". We will use an idea similar to relaxed planning graph (RPG) to optimize our search algorithm. Last but not least, we base our approach on some of the insights from LP-RPG \cite{cfls:08} about the number of occurrences of an action in a plan, but we will allow for more general terms like Boolean and linear repetitions.
	
	In general, many approaches deal with separate propositional variables from numeric ones. However, numeric variables can model propositional ones, as discussed before. Therefore, we will only use numeric variables and interpret the values as propositional ones, by $ 0 \mapsto \bot $, and $ 1 \mapsto \top $, while ensuring that only $ 0, 1 $ occur through the preconditions and effects.
	\section{Integer Addition Planning}
	The following definition will allow for a straightforward translation of numeric planning problems into an ILP, as we will show in Section 4. Note that we use \textit{registers} instead of \textit{variables} to avoid mixing the concept with ILP variables.
	\begin{definition}
		An \textit{integer addition action description} (denoted IAD) is a tuple $ \A = (A, R, \sigma, \pi) $, where $ A, R $ are non-empty sets of actions and registers respectively, $ \sigma : A \times R \rightarrow \Z $ defines the additive effect of each action on each register, $ \pi : A \times R \rightarrow \Z \cup \{ - \infty \} $ defines the preconditions of each action for each register.
	\end{definition}
	For each action $ b \in A $ and each register $ x \in R $, it is meant for the current value of register $ x $ to be greater than or equal to $ \pi(b, x) $ for $ b $ to be applicable. If $ \pi(b, x) = - \infty $, then $ b $ has no precondition w.r.t. register $ x $. After applying $ b $, the current value of register $ x $ is incremented by the (constant) integer value $ \sigma(b, x) $.
	\begin{definition}
		For an IAD $ \A = (A, R, \sigma, \pi) $, a \textit{situation} $ S : R \rightarrow \Z $ is a function mapping an integer value to each register. Additionally:
		\begin{itemize}
			\item For an action $ a \in A $, a situation \textit{satisfies} the preconditions of $ a $ iff. $ S(x) \geq \pi(a, x) $ for all $ x \in R $ (denoted $ S \vDash a $).
			\item An action $ a \in A $ can be \textit{added} to a situation by adding its effects, i.e., $ (S + a)(x) := S(x) + \sigma(a, x) $ for $ x \in R $.
		\end{itemize}
	\end{definition}
	Next, we will define the planning problem by defining the set containing its solutions.
	\begin{definition}
		For an IAD $ \A = (A, R, \sigma, \pi) $, an \textit{initial situation} $ S_{0} : R \rightarrow \Z $, and a \textit{goal action} $ g \in A $, an \textit{integer addition planning problem} (denoted IAP) $ \{ S_{0} \rightarrow g \}_{\A} $ is defined as the set of action lists $ l = [a_{0}, ..., a_{n}] \in A^{*} $, s.t. $ a_{n} = g $ and $ S_{i} \vDash a_{i} $, where $ S_{i} := S_{i-1} + a_{i-1} $ if $ i \geq 1 $, for all $ i \in \{0, .., n\} $, and for some $ n \in \N $.
	\end{definition}
	IAP is Turing-complete because it can model the so-called Abacus programs, i.e., it is undecidable, as shown by Helmert \shortcite{h:02}.
	\subsection{Offline Elevator Problem}
	Let us model an elevator with integers $ n $ as maximal and $ m $ as minimal floors. The elevator is initially at floor $ 0 $ and must deliver passengers given as tuples $ (f_{-}, f_{+}) \in \Z^2 $, meaning the passenger is at floor $ f_{-} $ and wants to go to floor $ f_{+} $. We want to find a plan that serves all passengers.
	\newcommand{\E}{\ensuremath{\mathcal{E}}}
	\begin{definition}
		An \textit{offline elevator problem} $ \E $ is a tuple $ (n, m, P) $, where $ n, m \in \Z $, s.t. $ m \leq 0 < n $, and $ P \subseteq \Z^2 $.
	\end{definition}
	\newcommand{\F}{\ensuremath{\mathbf{f}}}
	\newcommand{\IN}{\ensuremath{\mathbf{in}}}
	\newcommand{\OUT}{\ensuremath{\mathbf{out}}}
	\newcommand{\DLVD}{\ensuremath{\mathbf{dlvd}}}
	To model this problem in IAP, let us first discuss which registers are needed. We need a register for the current floor of the elevator $ \F $, and two registers for each passenger $ \IN(f_{-}, f_{+}), \DLVD(f_{-}, f_{+}) $, where $ \IN $ means the passenger is inside the elevator, and $ \DLVD $ means he has been delivered. Notice that $ \F$ is of linear nature, while $ \IN, \DLVD $ are of Boolean nature. Therefore, we use $ 0 \mapsto \bot; 1 \mapsto \top $ to interpret the values of these registers. The goal is to deliver all passengers. The actions needed for movement are $ u $, which takes us one floor up, and $ d $ in the other direction. We need two actions for each passenger (enter) $ e(f_{+}, f_{-}) $ that lets her in, and (leave) $ l(f_{+}, f_{-}) $ that lets her out. Let us use the notation shown in Figure \ref{fig:idem} to describe these actions as IAD, where $ a = (\text{preconditions of } a, \text{effects of } a) $:
	\begin{equation*} \begin{split}
			u := & ( \{ [\F \leq n - 1] \}, \{ [\F \peq 1] \}) \\
			d := & ( \{ [\F \geq m + 1 ] \}, \{ [ \F \meq 1 ] \}) \\
			e(f_{-}, f_{+}) := & ( \{[\F = f_{-}]\}, \{ [\IN(f_{-}, f_{+}) \peq 1] \} ) \\
			l(f_{-}, f_{+}) := & ( \{ [\F = f_{+}], [\IN(f_{-}, f_{+}) = 1] \}, \\
			& \, \, \{ [\IN(f_{-}, f_{+}) \meq 1], [\DLVD(f_{-}, f_{+}) \peq 1] \} ) \\
			g := & ( \{ [\DLVD(f_{-}, f_{+}) = 1] : (f_{-}, f_{+}) \in P \}, \{ \} )
	\end{split} \end{equation*}
	
	This means that for a an offline elevator instance denoted $ \E_{i} = (n_{i}, m_{i}, P_{i}) $ we can define an IAD instance $ \A_{i} = (A_{i}, R_{i}, \sigma_{i}, \pi_{i}) $, where, e.g., $ \sigma_{i}(u, \F) = 1 $ for $ [\F \peq 1] $ and $ \sigma_{i}(u, \IN(f_{-}, f_{+})) = 0 $, because $ u $ has no effect on $ \IN(f_{-}, f_{+}) $. \\
	Additionally, $ \loprec_{i}(e(f_{-}, f_{+}), \F) = \upprec_{i}(e(f_{-}, f_{+}), \F) = f_{-} $ for $ [\F = f_{-}] $ (i.e., $ [\F \geq f_{-}] $ and $ [\F \leq f_{-}] $), and so on. \\
	The initial situation is defined by $ S_{0}(x) := 0 $ for all $ x \in R_{i} $, for $ R_{i} := \{ \F, \IN(f_{-}, f_{+}), \DLVD(f_{-}, f_{+}) : (f_{-}, f_{+}) \in P_{i} \} $.
	
	We will use the offline elevator problem as a running example (always in \textit{italic font}) demonstrating the concepts and algorithms discussed in the paper.
	\begin{example}
		Let $ \E_{1} = (0, 3, \{ (3, 1) \}) $, i.e. there are four floors $ \{0, 1, 2, 3\} $, and one passenger who waits in floor $ 3 $ and wants to go to floor $ 1 $. Then, the minimal solution to $ \E_{1} $ is $ l_{1} = [u, u, u, e(3, 1), d, d, l(3, 1), g] $.
	\end{example}
	
	\section{Multi-Valued Partial Order Plans}
\begin{figure}
		\centering
		\includegraphics[scale=0.09]{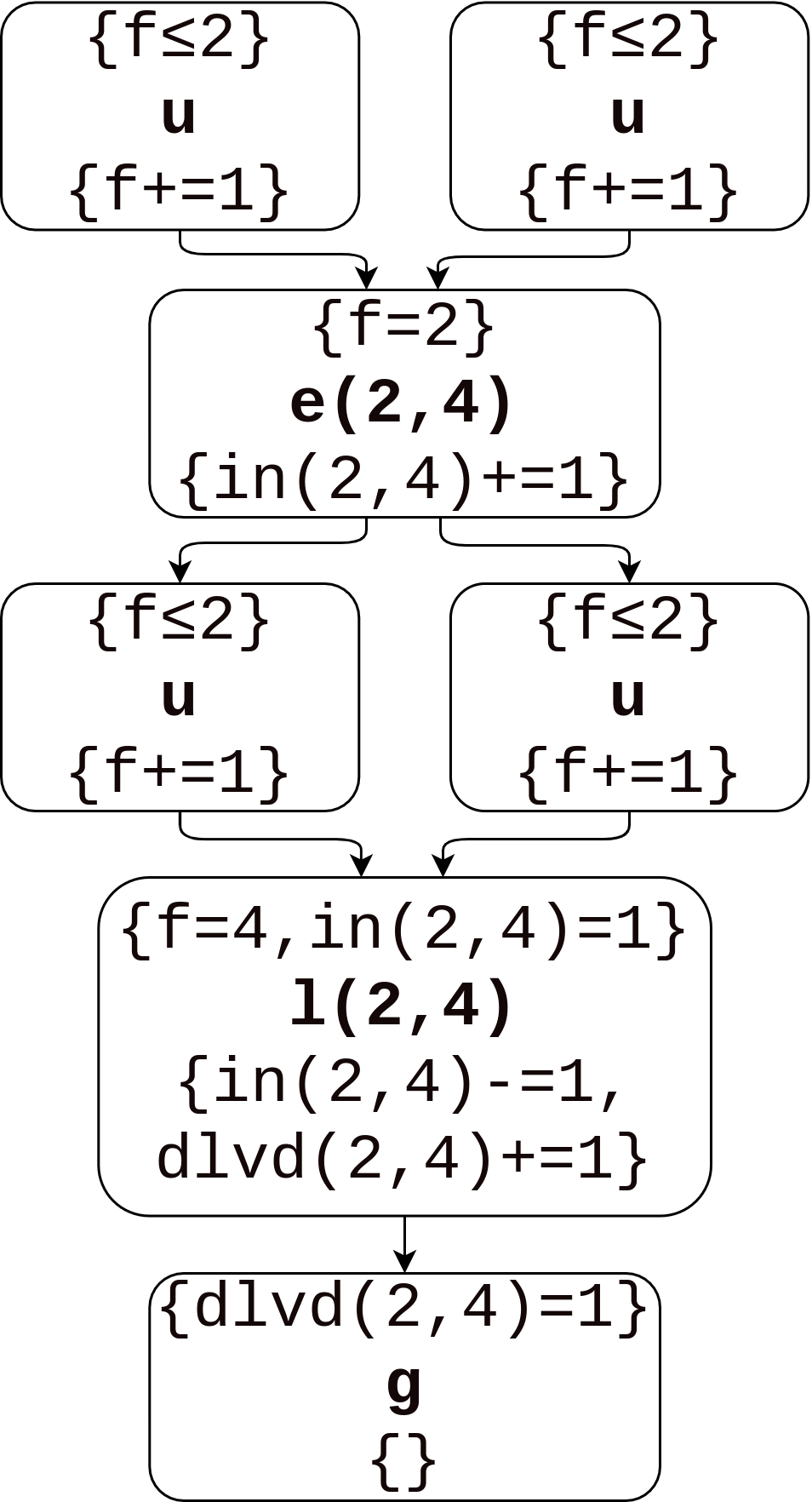}
		\caption{Minimal partial order representing $ l_{2} $ from Ex. 2. Notice the need for two ordered copies of $ u $.}
		\label{fig:e2}
	\end{figure}

	Let us first discuss the idea of a multi-valued relation. For example, a classical binary relation $ r $ over a set $ K $ is a subset of $ K^2 $. We can represent this relation by $ f_{r} : K^2 \rightarrow \{ 0,1 \} $, s.t. $ f_{r}(u, e) = 1 $ iff. $ (u, e) \in r $. Here, we extend the concept by allowing for values other than $ 0 $ and $ 1 $, e.g., $ f_{r} : K^2 \rightarrow \N $. In planning, we usually represent solutions as a partial order relation $ < $. An action $ u $ is ordered before $ e $, denoted $ u < e $, here $ f_{<}(u, e) = 1 $, if $ u $ must happen before $ e $. With $ f_{<}(u, e) = 2 $ can represent that $ u $ must happen twice before $ e $. We will use this representation for the number of needed \textit{linear} repetitions of actions. However, $ u $ might need to happen twice before $ e $ and twice after it.
	\begin{example}
		The instance $ \E_{2} = (0, 4, \{(2, 4)\}) $, with the solution $ l_{2} = [u, u, e(2, 4), u, u, l(2, 4), g] $. In this solution, two occurrences of $ u $ are needed before and after one occurrence of $ e(2, 4) $. Compare Figure \ref{fig:e2}.
	\end{example}
	We cannot represent it by $ f_{<}(u, e) = 2 $, $ f_{<}(e, u) = 1 $, because this creates a cyclic plan. Let $ u' \neq u $ be a new action with the same preconditions and effects as $ u $. The plan can be represented as $ f_{<}(u, e) = 2 $, $ f_{<}(e, u') = 2 $, where $ u $ and $ u' $ are different actions but have the same preconditions and effects. Here, $ u' $ is a \textit{Boolean} repetition of $ u $. Notice that Boolean repetitions could also work as linear ones since the plan $ [u, u'] $ is equivalent to the plan $ [u', u] $; thus, we need to search for a minimal number of Boolean repetitions needed to maintain \textit{compactness} of the solution representation.
	
	In general, all linear repetitions of an action do not have to happen in any specific order, as the effects are added together at the end, and addition is associative. However, Boolean repetitions require a specific order. For this reason, we will allow for two kinds of action repetitions: \textit{Unordered} and \textit{ordered} copies, which will serve the purposes of linear and Boolean repetitions, respectively. 
	
	Technically speaking, we will use multi-sets of actions $ \Aa = (A, \mu) $ for $ \mu : A \rightarrow \N $, where $ \mu(a) $ is equal to the number of \textit{ordered copies} of action $ a $. Additionally, we will use matrices $ \Pa \in \N^{|\Aa| \times |\Aa|} $ to describe the multi-valued relation $ f_{<} $ mentioned before. $ \Pa[b, a] $ stands for the entry in the row of $ b $ and column of $ a $ in the matrix $ \Pa $, and it represents the number of \textit{unordered copies} of $ a $ before $ b $, i.e. $ f_{<}(a, b) $.
	
	\textit{In Example 2, $ l_{2} = [u, u, e(2, 4), u, u, l(2, 4), g] $ (Fig.  \ref{fig:e2}). Let $ \Aa_{2} $ be the multi-set $ \{g, l(2, 4), u, e(2, 4), u\} $, then, the MvPOP $ \Pa_{2} $ over $ \Aa_{2} $ with max. $ g $ described by following matrix, is an equivalent representation to $ l_{2} $:
		\begin{center}
			\small
			\begin{tabular}{c||c|c|c|c|c}
				$ \Pa_{2} $ & $ g $ & $ l(2, 4) $ & $ u $ & $ e(2, 4) $ & $ u $ \\
				\hline \hline
				$ g $ & $ 0 $ & $ 1 $ & $ 2 $ & $ 1 $ & $ 2 $ \\
				\hline
				$ l(2, 4) $ & $ 0 $ & $ 0 $ & $ 2 $ & $ 1 $ & $ 2 $ \\
				\hline
				$ u $ & $ 0 $ & $ 0 $ & $ 0 $ & $ 1 $ & $ 2 $ \\
				\hline
				$ e(2, 4) $ & $ 0 $ & $ 0 $ & $ 0 $ & $ 0 $ & $ 2 $ \\
				\hline
				$ u $ & $ 0 $ & $ 0 $ & $ 0 $ & $ 0 $ & $ 0 $
			\end{tabular}
		\end{center}
		Notice that are two different rows and columns for $ u $. For example, $ \Pa_{2}[e(2, 4), u] = 2 $(4th row, 5th column) corresponds to the two occurrences of the first ordered copy of $ u $ before $ e(2, 4) $, and $ \Pa_{2}[u, e(2, 4)] = 1 $(3rd row, 4th column) corresponds for the single occurrence of $ e(2, 4) $ before the second ordered copy of $ u $.}
	
	In general, there will be multiple but different rows and columns for an action $ a $ if $ \mu(a) > 1 $. To keep the notations simple, we mean all rows/columns by \textit{for all} $ a \in \Aa $, and one row/column by \textit{there exists} $ a \in \Aa $.
	\begin{definition}
		An MvPOP $ \Pa $ over a multi-set $ \Aa = (A, \mu) $ is a matrix in $ \N^{|\Aa| \times |\Aa|} $ s.t. for all $ a, b, c \in \Aa $:
		\begin{itemize}
			\item Irreflexive: $  \Pa[a, a] = 0 $.
			\item Asymmetric: $ \Pa[b, a] > 0 $ implies $ \Pa[a, b] = 0 $.
			\item Transitive: $ \Pa[c, b] > 0 $ implies $ \Pa[c, a] \geq \Pa[b, a] $.
		\end{itemize}
		We say that $ g \in \Aa = (A, \mu) $ is the maximum of $ \Pa $ iff. $ \mu(g) = 1 $ and $ g $ is the only action with $ \Pa[a, g] = 0 $ for all actions $ a \in \Aa \setminus \{g\} $.
	\end{definition}
	Classically, partial order plans stand for a set of action lists, known as \textit{linearizations} or \textit{linear extensions} of the partial order (which has nothing to do with linear repetitions discussed in this paper). For example, if $ u < e $, then $ u $ should occur before $ e $ in any linearization of $ < $. However, in an MvPOP $ \Pa $, there might be multiple occurrences of $ u \in \Aa $. For this reason, we will assume that a linearization of $ \Pa $ must have $ \Pa[b, a] $ occurrences of $ a $ before the first occurrence of $ b $, for all $ a, b \in \Aa $.
	
	For any action list $ l \in A^* $, let $ l[..b] \in A^* $ be the sub-list of $ l $ from the first action to the first occurrence of $ b $. For example, if $ l = [a, c, c, b, e, b, g] $, then $ l[..b] = [a, c, c, b] $. Additionally, let $ \#_{a}(l) $ be the number of occurrences of action $ a $ in $ l $. Notice that in any MvPOP with max. $ g $, the number of occurrences of an action $ a \in \Aa $ should be fixed $ \#_{a}(l) = \Pa[g, a] $ for all linearizations $ l $ of $ \Pa $, similar to linearizations of partial orders. Because of transitivity, and because $ g $ is the maximum, i.e., no actions occur after it in all linearizations $ l $ of $ \Pa $, then $ l[..b] $ must be contained in $ l[..g] $, i.e., $ \Pa[g, a] \geq \Pa[b, a] $. Therefore, the number of occurrences of $ a $ before $ b $ must be in any linearization between $ \Pa[b, a] $ and $ \Pa[g, a] $, for all $ a, b \in \Aa $.
	\begin{definition}
		The set of \textit{linearizations} of an MvPOP $ \Pa $ (denoted $ \lin(\Pa) $) over $ \Aa = (A, \mu) $ with max. $ g $ is the set of action lists $ l \in A^{*} $ s.t. $ \Pa[b, a] \leq  \#_{a}(l[..b]) \leq \Pa[g, a] $ for all $ a, b \in \Aa $.
	\end{definition}
	\begin{example}
		$ \E_{3} = (0, 2, \{(0, 2), (2, 1)\}) $, with two solutions $ l^1_{3} = [e(0, 2), u, u, l(0, 2), e(2, 1), d, l(2,1), g] $, and $ l^2_{3} = [e(0, 2), u, u, e(2, 1), l(0, 2), d, l(2,1), g] $. However, one MvPOP $ \Pa_{3} $ can represent both linearizations over $ \{g, l(2, 1), d, e(2, 1), l(0, 2), u, e(0,2)\} $, by:
		\begin{center}
			\small
			\begin{tabular}{c||c|c|c|c|c|c|c}
				$ \Pa_{3} $     & $ g $ & $ l(2,1) $ & $ d $ & $ e(2, 1) $ & $ l(0, 2) $ & $ u $ & $ e(0, 2) $ \\ \hline \hline
				$ g $       & $ 0 $     & $ 1 $          & $ 1 $     & $ 1 $           & $ 1 $           & $ 2 $     & $ 1 $           \\ \hline
				$ l(2,1) $  & $ 0 $     & $ 0 $          & $ 1 $     & $ 1 $           & $ 1 $           & $ 2 $     & $ 1 $           \\ \hline
				$ d $       & $ 0 $     & $ 0 $          & $ 0 $     & $ 1 $           & $ 1 $           & $ 2 $     & $ 1 $           \\ \hline
				$ e(2, 1) $ & $ 0 $     & $ 0 $          & $ 0 $     & $ 0 $ & $ \mathbf{0} $ & $ 2 $     & $ 1 $           \\ \hline
				$ l(0, 2) $ & $ 0 $     & $ 0 $          & $ 0 $     & $ \mathbf{0} $ & $ 0 $ & $ 2 $     & $ 1 $           \\ \hline
				$ u $       & $ 0 $     & $ 0 $          & $ 0 $     & $ 0 $              & $ 0 $              & $ 0 $     & $ 1 $           \\ \hline
				$ e(0, 2) $ & $ 0 $     & $ 0 $          & $ 0 $     & $ 0 $              & $ 0 $              & $ 0 $     & $ 0 $           \\
			\end{tabular}
		\end{center}
		The two marked $ \mathbf{0} $'s represent that both $ e(2, 1), l(0,2) $ are incomparable to each other. More on that later.
	\end{example}
	
	\subsection{Single-Linearization MvPOPs}
	Let us first focus on MvPOPs that have exactly one linearization to understand how they relate to action lists.
	\begin{lemma}
		\label{lem:lin0}
		For any action list $ l = [a_{0}, ..., a_{n}] \in A^{*} $ there exists a multi-set $ \Aa = (A, \mu) $ and an MvPOP $ \Pa_{l} $ over $ \Aa $ with max. $ a_n $ s.t. $ \lin(\Pa_{l}) = \{l\} $.
	\end{lemma}
	\begin{proof}
		Let $ \mu(a) $ be the number of occurrences of $ a $ in $ l $. Additionally, let $ \Pa_{l}[b, a] = 1 $ if $ a $ occurs before $ b $ in $ l $ and $ \Pa_{l}[b, a] = 0 $ else, for $ a, b \in \Aa $. Then, the only linearization possible of $ \Pa_{l} $ is $ l $.
	\end{proof}
	\begin{example}
		$ \E_{4} = (0, 2, \{(0,2)\}) $, with a given solution $ l_{4} = [e(0,2), u, u, l(0,2), g] $. We describe the MvPOP $\Pa_{4} $ over $ \Aa_{4} := \{g, l(0, 2), u, u, e(0,2) \} $ s.t. $ \lin(\Pa_{4}) = \{l_{4}\} $:
		\begin{center}
			\small
			\begin{tabular}{c||c|c|c|c|c}
				$ \Pa_{4} $    & $ g $ & $ l(0,2) $ & $ u $ & $ u $ & $ e(0,2) $ \\ \hline \hline
				$ g $      & $ 0 $     & $ 1 $          & $ 1 $     & $ 1 $     & $ 1 $          \\ \hline
				$ l(0,2) $ & $ 0 $     & $ 0 $          & $ 1 $     & $ 1 $     & $ 1 $          \\ \hline
				$ u $      & $ 0 $     & $ 0 $          & $ 0 $     & $ 1 $     & $ 1 $          \\ \hline
				$ u $      & $ 0 $     & $ 0 $          & $ 0 $     & $ 0 $     & $ 1 $          \\ \hline
				$ e(0,2) $ & $ 0 $     & $ 0 $          & $ 0 $     & $ 0 $     & $ 0 $       
			\end{tabular}
		\end{center}
	\end{example}	
	Now, let us represent the IAP problem as a set of linear inequalities. For a multi-set $ \Aa = (A, \mu) $, we will represent the IAD $ \A = (A, R, \sigma, \loprec, \upprec) $ with matrices $ \Sigma, \Loprec, \Upprec $ in $ (\Z \cup \{ \infty, - \infty \})^{|\Aa| \times |R|} $, where we define $ \Sigma[a, x] := \sigma(a, x) $, $ \Pi^\circ[a, x] := \pi^\circ(a, x) $, for $ \circ \in \{\geq,\leq\} $, $ a \in \Aa $, and $ x \in R $. Additionally, the matrix $ \Ss_{0} \in \Z^{|\Aa| \times |R|}  $ represents an initial situation $ S_{0} $, by $ \Ss_{0}[a, x] = S_{0}(x) $ for $ a \in \Aa, x \in R $.
	
	\textit{In Example 4, $ \Sigma_{4} $ over $ \Aa_{4} $ is:
		\begin{center}
			\small
			\begin{tabular}{c||c|c|c}
				$ \Sigma_{4} $ & $ \F $ & $ \IN(0, 2) $ & $ \DLVD(0, 2) $ \\ \hline \hline
				$ g $       & $ 0 $      & $ 0 $             & $ 0 $               \\ \hline
				$ l(0, 2) $ & $ 0 $      & $-1 $             & $ 1 $               \\ \hline
				$ u $       & $ 1 $      & $ 0 $             & $ 0 $               \\ \hline
				$ u $       & $ 1 $      & $ 0 $             & $ 0 $               \\ \hline
				$ e(0, 2) $ & $ 0 $      & $ 1 $             & $ 0 $           
			\end{tabular}
		\end{center}
		Notice the two identical rows for $ u $, which represent the two ordered copies of $ u $. Now, if $ \Ss := \Ss_{0} + \Pa_{4} \Sigma_{4} $, then
		\begin{equation*}
			\begin{split}
				\Ss[e(0,2), \F] = & 0 \geq \Loprec[e(0,2), \F] \\
				\Ss[l(0,2), \F] = & 1 \cdot \sigma(u, \F) + 1 \cdot \sigma(u, \F) + 1 \cdot \sigma(e(0, 2), \F) \\
				= & 1 + 1 + 0 = 2 \leq \Upprec[l(0,2), \F]
			\end{split} 
		\end{equation*} 
		i.e., $ \Ss[e(0,2), \F], \Ss[l(0,2), \F] $ are the values of register $ \F $ before applying the actions $ e(0,2) $, $ l(0, 2) $ respectively.}
	
	\begin{lemma}
		\label{lem:lin}
		In IAP $ \{ S_{0} \rightarrow g \}_{\A} $, if $ l \in \{ S_{0} \rightarrow g \}_{\A} $, then:
		$$	\Loprec \leq \Ss_{0} + \Pa_{l} \Sigma \leq \Upprec $$
		where $ \leq $ is applied element-wise.
	\end{lemma}
	\begin{proof}
		Let $ \Ss := \Ss_{0} + \Pa_{l} \Sigma $, i.e.: $$ \Ss[b, x] = S_{0}(x) + \sum_{a \in \Aa} \Pa_{l}[b, a] \sigma(a, x) $$
		Then, $ \Ss[b, x] $ represents the situation of register $ x \in R $ before one of the occurrences of action $ b $ in $ l $, because $ \Pa_{l}[b, a] = 1 $ iff. $ a $ is before $ b $ in $ l $. This holds for all occurrences of $ b $ in $ l $ because there is a row in $ \Ss $ for each such occurrence. Since $ \Loprec \leq \Ss_{0} + \Pa_{l} \Sigma \leq \Upprec $, then $ \loprec(b, x) \leq \Ss[b, x] \leq \upprec(b, x) $ for all $ b $ occurring in $ l $ and all registers $ x \in R $.
	\end{proof}
	\subsection{Incomparability and Violation in MvPOPs}
	In partial order plans, two actions that are unordered (incomparable) to each other can cause a \textit{threat} or a \textit{violation}. This is solved classically by demotion or promotion. A similar phenomenon arises in MvPOPs, but the number of these incomparable occurrences here matters. In the single-linearization case, the set $ \lin(\Pa) $ was a singleton, which did not lead to any complications. However, if $ \lin(\Pa) $ contains at least two different action lists, then at least two occurrences of two actions $ a, b \in \Aa $ are ordered differently in these lists. We call such occurrences incomparable.
	\begin{example}
		$ \E_{5} = (0, 3, \{(1, 3)\}) $, and the MvPOP $ \Pa_{5} $ over $ \{ g, l(1, 3), e(1, 3), u \} $:
		\begin{center}
			\small
			\begin{tabular}{c||c|c|c|c}
				$ \Pa_{5} $     & $ g $ & $ l(1,3) $ & $ e(1, 3) $ & $ u $          \\ \hline \hline
				$ g $       & $ 0 $     & $ 1 $          & $ 1 $           & $ \mathbf{3} $              \\ \hline
				$ l(1,3) $  & $ 0 $     & $ 0 $          & $ 1 $           & $ 3 $ \\ \hline
				$ e(1,3) $ & $ 0 $     & $ 0 $          & $ 0 $           & $ \mathbf{1} $ \\ \hline
				$ u $       & $ 0 $     & $ 0 $          & $ 0 $           & $ 0 $            
			\end{tabular}
		\end{center}
		Notice the marked numbers $ \mathbf{3}, \mathbf{1} $, i.e., that the number of occurrences of $ u $ before $ e(1, 3) $ varies between $ 1 $ and $ 3 $; thus:
		\begin{equation*}
			\begin{split}
				\lin(\Pa_{5}) = & \{ l_{5}^1, l_{5}^2, l_{5}^3 \} \text{, where} \\
				l_{5}^1 = & [u, u, u, e(1, 3), l(1, 3), g] \\
				l_{5}^2 = & [u, u, e(1, 3), u, l(1, 3), g] \\
				l_{5}^3 = & [u, e(1, 3), u, u, l(1, 3), g]
			\end{split}
		\end{equation*}
		\begin{figure}
			\centering
			\includegraphics[scale=0.09]{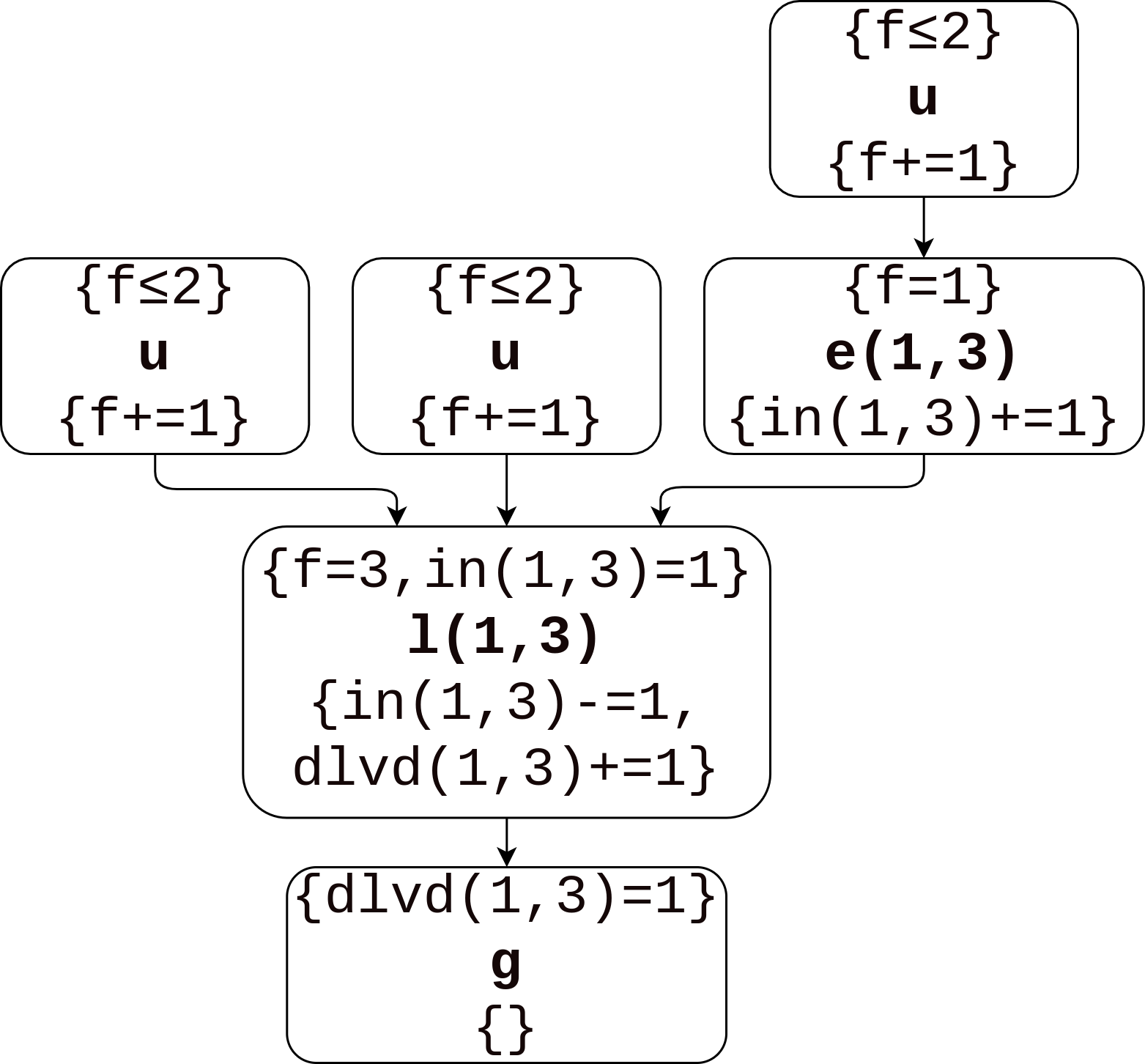}
			\caption{A partial order constructed from the three actions lists in $ \lin(\Pa_{5}) = \{ l_{5}^1, l_{5}^2, l_{5}^3 \} $ of Example 5. Notice the two  incomparable occurrences of $ u $ to $ e(1, 3) $.}
			\label{fig:e5}
		\end{figure}
	\end{example}
	\begin{definition}
		\label{def:inc0}
		For any MvPOP $ \Pa $ over $ \Aa $ with max. $ g $ and an action $ b \in \Aa $, we say that there are $ n $ \textit{incomparable occurrences} of $ a $ to $ b $ in $ \Pa $ (denoted $ \I^{\Pa}[b,a] = n $) iff. there are at least $ n $ occurrences of $ a $ in any linearization of $ \Pa $ (i.e., $ \Pa[g, a] \geq n $), and there are $ n + 1 $ linearizations s.t. $ a $'s occurrences are split into $ i $ before $ b $ and $ n - i $ after $ b $ for all $ i \in \{0, ..., n\} $.
	\end{definition}
	\begin{lemma}
		\label{lem:inc}
		In any MvPOP $ \Pa $ over $ \Aa $ with max. $ g $ and actions $ a, b \in \Aa $, s.t. $ a \neq g $, and $ \Pa[g, b] > 0 $:
		\begin{itemize}
			\item $ \I^{\Pa}[b, a] = \Pa[g, a] - \Pa[b, a] $ if $ a \neq b $.
			\item $ \I^{\Pa}[b, b] = \Pa[g, b] - 1 $.
		\end{itemize}
	\end{lemma}
	\begin{proof}
		First, notice that $ \Pa[g, a] - \Pa[b, a] \geq 0 $. In any linearization, there are $ \Pa[g, a] $ occurrences of $ a $. $ \Pa[b, a] $ represents the number of $ a $'s occurrences that must come before $ b $; thus $ \Pa[g, a] - \Pa[b, a] $ are free to occur before or after $ b $. Additionally, there are $ \Pa[g, a] - 1 $ occurrences of $ a $ that are incomparable to each of all the occurrences of $ a \in \Aa $.
	\end{proof}
	\textit{In Example 5, $ \I^{\Pa_{5}}[u, u] = 2 $, $ \I^{\Pa_{5}}[e(1, 3), u] = 2 $. For this reason, we had three linearizations $ l_{5}^1, l_{5}^2, l_{5}^3 $. Now, notice that $ l_{5}^3 $ is a valid solution of $ \E_{5} $, but both $ l_{5}^1 $ and $ l_{5}^2 $ are not, because $ e(1, 3) $ has the precondition $ [\F = 1] $, which stands for an upper and a lower bound w.r.t. register $ \F $, i.e., $ \loprec(e(1, 3), \F) = \upprec(e(1, 3), \F) = 1 $. The upper bound gets violated in $ l_{5}^1, l_{5}^2 $, because $ u $ has the effect $ [\F \peq 1] $. In other words, the pair $ (u, e(1, 3)) $ cause a threat, compare Fig. \ref{fig:e5}.}
	\begin{definition}
		For an IAD $ \A = (A, R, \sigma, \loprec, \upprec) $, an action $ a $ \textit{violates} the lower bound of $ b $ w.r.t. register $ x $ ($ a \viogeq_{x} b $) iff. there exists a situation $ S : R \rightarrow \Z $, s.t. $ S(x) \geq \loprec(b, x) $, but $ (S + a)(x) < \loprec(b, x) $. Similarly for the upper bound $ a \violeq_{x} b $, if $ S(x) \leq \upprec(b, x) $, but $ (S + a)(x) > \upprec(b, x) $. In general, $ a \sim b $ iff. there is $ x \in R \stt a \violeq_{x} b \ort a \viogeq_{x} b $.
	\end{definition}
	
	\textit{In Example 5, if $ S(\F) = 1 \leq 1 = \upprec(e(1, 3), \F) $, then, $ (S + u)(\F) = 2 > \upprec(e(1, 3), \F) $, i.e., $ u \violeq_{\F} e(1, 3) $.}
	
	\begin{definition}
		\label{def:inc}
		For an IAP $ \{ S_{0} \rightarrow g \}_{\A} $, the \textit{violation multi-valued relations} of MvPOP $ \Pa $ over $ \Aa $ with max. $ g $ w.r.t. register $ x \in R $ (denoted $ \I_{\leq x}^{\Pa}, \I_{\geq x}^{\Pa} \in \N^{|\Aa| \times |\Aa|} $) is defined as:
		\begin{itemize}
			\item $ \I_{\circ x}^{\Pa}[b, a] := \I^{\Pa}[b, a] $ if $ a \viocirc_{x} b $ and $ b \neq g $.
			\item $ \I_{\circ x}^{\Pa}[b, a] := 0 $ else.
		\end{itemize} 
		where  $ a, b \in \Aa $, and $ \circ $ is a place holder for either $ \leq $ or $ \geq $.
	\end{definition}
	\textit{In Example 5, notice that $ \Pa_{5} + \I_{\leq\F}^{\Pa_{5}} $ represents the worst situation for the precondition $ [\F = 1] $ of $ e(1, 3) $. Notice the marked $ \mathbf{3} $ and compare Fig. \ref{fig:e5}: 
		\begin{center}
			\small
			\begin{tabular}{c||c|c|c|c}
				$ \Pa_{5} + \I_{\leq \F}^{\Pa_{5}} $     & $ g $ & $ l(1,3) $ & $ e(1, 3) $ & $ u $          \\ \hline \hline
				$ g $       & $ 0 $     & $ 1  $         & $ 1  $           & $ 3 $           \\ \hline
				$ l(1,3) $  & $ 0 $     & $ 0 $          & $ 1  $           & $ 3 $ \\ \hline
				$ e(1,3) $ & $ 0 $     & $ 0 $         &$ 0 $           & $ \mathbf{3} $ \\ \hline
				$ u $       & $ 0 $     & $ 0 $          & $ 0 $           & $ 2 $            
			\end{tabular}
	\end{center}}
	Based on these insights, we can formulate any IAP as an ILP instance by checking for violations using the incomparable multi-valued relation defined above.
	\begin{theorem}
		\label{th:nonlin}
		For an IAP $ \{ S_{0} \rightarrow g \}_{\A} $, a multi-set $ (A, \mu) $, and an MvPOP $ \Pa $ over $ (A, \mu) $ with max. $ g $: 
		\begin{equation} \label{eq:inc} \begin{split}
				\text{If} \forallt x \in R \andt & \forallt \circ \in \{ \geq, \leq \} : \\ & \Ss_{0} + (\Pa + \I^{\Pa}_{\circ x}) \Sigma \: \circ \: \Pi^\circ \\
				\text{then, } & \lin(\Pa) \subseteq \{ S_{0} \rightarrow g \}_{\A}
		\end{split} \end{equation}
	\end{theorem}
	\begin{proof}
		Let $ \Ss^+ := \Ss_{0} + \Pa \Sigma $. Similar to the proof of Lemma \ref{lem:lin}, we see that in any linearization $ l $ where all actions incomparable to $ b $ occur after $ b $, the $ x $-bounds of $ b $ are satisfied, because $ \loprec(b, x) \leq \Ss^+[b, x] \leq \upprec(b, x) $. However, we need to ensure that the $ x $-bounds of $ b $ are satisfied for linearizations where part of the incomparable actions to $ b $ occur before $ b $. \\
		Let $ \Ss := \Ss_{0} + (\Pa + \I_{\circ x}^{\Pa}) \Sigma = \Ss^+ + \I_{\circ x}^{\Pa} \Sigma $. I.e.: \begin{equation} \label{eq:linspec} 
			\Ss[b, x] = \Ss^+[b, x] + \sum_{a \in \Aa} \I_{\circ x}^{\Pa}[b, a] \sigma(a, x) \end{equation}
		There are two cases for $ \I_{\circ x}^{\Pa}[b, a] $ as defined in Def. \ref{def:inc}:
		\begin{itemize}
			\item If $ a \not \viocirc_{x} b $, then, none of the incomparable occurrences of $ a $ can invalidate the $x$-bounds of $ b $ if $ \Ss^+[b, x] $ is already within these bounds. This is the case where $ \I_{\circ x}^{\Pa}[b, a] = 0 $.
			\item If $ a \viocirc_{x} b $, then we need to make sure that the worst-case linearization, i.e., where all incomparable $ a $'s occur before $ b $, still does not invalidate the $ x $-bounds of action $ b $. The value of $ \I_{\circ x}^{\Pa}[b, a] $, in this case, is the maximal number of incomparable $ a $'s occurrences to $ b $, compare Lemma \ref{lem:inc}.
		\end{itemize}
		From (\ref{eq:inc}), (\ref{eq:linspec}), we know that
		\begin{equation*}
			\begin{split}
				& \Ss^+[b, x] + \sum_{a \in \Aa} \I_{\geq x}^{\Pa}[b, a] \sigma(a, x) \geq \loprec(b, x) \\
				\andt & \Ss^+[b, x] + \sum_{a \in \Aa} \I_{\leq x}^{\Pa}[b, a] \sigma(a, x) \leq \upprec(b, x)
			\end{split}
		\end{equation*}
		Therefore, even the worst-case linearizations w.r.t. register $ x \in R $, i.e., where the effects of all violating, incomparable actions $ \sum_{a \in \Aa} \I_{\circ x}^{\Pa}[b, a] \sigma(a, x) $ are counted before $ b $, the $ x $-bounds of $ b $ still hold. Since (\ref{eq:inc}) is quantified over all registers $ x \in R $ for both directions $ \circ \in  \{\leq, \geq \} $, thus, any linearization $ l \in \lin(\Pa) $ is in $ \{ S_{0} \rightarrow g \}_{\A} $.
	\end{proof}
	\textit{In Example 5, there exists an MvPOP that represents only the valid solution $ l_{5}^3 $, but it needs at least two ordered copies of $ u $, i.e., $ \mu(u) \geq 2 $. In other words, no MvPOP over $ (A_{5}, \mu) $ can solve the inequality in (\ref{eq:inc}) for $ \E_{5} $, if $ \mu(u) < 2 $.}
	\section{IAP as Search}
	In this section, we use Th. \ref{th:nonlin} to reformulate the planning problem as a combination of ILP and search problems.
	\begin{definition}
		For an IAP $ \{ S_{0} \rightarrow g \}_{\A} $, an \textit{integer addition planning fixed problem} $ [ S_{0} \rightarrow g ]_{\Aa} $ is defined as the set of linearizations of the MvPOPs $ \Pa $ over a fixed multi-set $ \Aa $ with max. $ g $ s.t. for all $ x \in R $:
		$$
		\Forallt \circ \in \{ \leq, \geq \} : \Ss_{0} + (\Pa + \I^{\Pa}_{\circ x}) \Sigma \: \circ \: \Pi^\circ
		$$
	\end{definition}
	Note that $ [S_{0} \rightarrow g]_{\Aa} $ is an ILP-feasibility problem.
	\begin{lemma}
		\label{lem:subset}
		For any IAP $ \{ S_{0} \rightarrow g \}_{\A} $ and any multi-sets $ \Aa_{1} = (A, \mu_{1}) $ and $ \Aa_{2} = (A, \mu_{2}) $, if $ \mu_{1}(a) \leq \mu_{2}(a) $ for all $ a \in A $, then:
		$$ [S_{0} \rightarrow g]_{\Aa_{1}} \subseteq [S_{0} \rightarrow g]_{\Aa_{2}} $$
	\end{lemma}
	\begin{proof}
		If there exists an action $ a \in A $, s.t. $ \mu_{2}(a) > \mu_{1}(a)$, then an MvPOP $ \Pa $ over $ \Aa_{2} $ allows for further distinction between occurrences of action $ a $, i.e., there will be extra rows and columns for $ a $ in $ \Pa $, which can be used if more ordered copies of $ a $ are needed. In other words, an MvPOP over $ \Aa_{2} $ has more expressive power than an MvPOP over $ \Aa_{1} $.
	\end{proof}
	
	\begin{corollary}[IAP as Search]
		\label{cor:iaps}
		For any IAP $ \{ S_{0} \rightarrow g \}_{\A} $:
		$$ \{S_{0} \rightarrow g\}_{\A} = \bigcup_{\substack{\Aa = (A, \mu) \\ \mu : A \rightarrow \N}} [S_{0} \rightarrow g]_{\Aa} $$
		$$ \{S_{0} \rightarrow g\}_{\A} = \lim_{n \rightarrow \infty} [S_{0} \rightarrow g]_{(A, n)} $$
	\end{corollary}
	\begin{proof}
		From Lemmata \ref{lem:lin0}, \ref{lem:lin}: $$ \{S_{0} \rightarrow g\}_{\A} \subseteq \bigcup_{\substack{\Aa = (A, \mu) \\ \mu : A \rightarrow \N}} [S_{0} \rightarrow g]_{\Aa} $$
		And from Theorem \ref{th:nonlin}: $$ \bigcup_{\substack{\Aa = (A, \mu) \\ \mu : A \rightarrow \N}} [S_{0} \rightarrow g]_{\Aa} \subseteq \{S_{0} \rightarrow g\}_{\A} $$
		By $ (A, n) $ we mean that $ \mu(a) = n $ for all $ a \in A $. \\
		From Lemma \ref{lem:subset}:
		$$ \{S_{0} \rightarrow g\}_{\A} = \lim_{n \rightarrow \infty} [S_{0} \rightarrow g]_{(A, n)} $$
	\end{proof}
	
	\begin{algorithm}
		\caption{IAP as Search}
		\KwIn{IAP instance $ \{ S_{0} \rightarrow g \}_{\A} $.}
		\KwOut{Action list $ l \in \{ S_{0} \rightarrow g \}_{\A} $.}
		$ \mu(a) \leftarrow 1 $ for each $ a \in A $; \\
		\While{True}{
			$ \Aa \leftarrow (A, \mu) $;
			\begin{equation*}
				\begin{split} \text{Minimize } f(\Pa) \stt & \forall x \in R \, \forall \circ \in \{\leq, \geq \}: \\ &  \Ss_{0} + (\Pa + \I^{\Pa}_{\circ x}) \Sigma \:  \circ \: \Pi^\circ
				\end{split}
			\end{equation*}
			\uIf{$ \exists \Pa $ MvPOP solution over $ \Aa $ with max. $ g $}{\Return any $ l \in \lin(\Pa) $;}
			\uElseIf{$ \Phi(\Aa) \neq \emptyset $}{$ \mu(a) \leftarrow \mu(a) + 1 $ for each $ a \in \Phi(\Aa) $;}
			\Else{\Return ``No solution found";}
		}
		\label{alg:rand}
	\end{algorithm}
	Corollary \ref{cor:iaps} shows how IAP can be formulated as a search problem that aims to find the (minimal) number of needed Boolean repetitions of each action, i.e., a minimal multi-set $ \Aa $. We can easily calculate the number of unordered copies (linear repetitions) needed of each action $ a \in A $ because we can calculate that using ILP, as shown in Th. \ref{th:nonlin}, which is a decidable problem. Therefore, the undecidability of IAP comes from the hardness of finding the needed number of ordered copies (Boolean repetitions) in the plan (i.e., $ \mu(a) $ for each action $ a $), specifically, knowing whether such a number exists or not.
	
	Algorithm \ref{alg:rand} is presented in a general way, where the functions $ \Phi $ and $ f $ are left undefined. We will refer to any special definitions of them in \textit{specifications}. We will specify $ f(\Pa) $ in Section 8. For now, let $ f(\Pa) := 0 $ independent of $ \Pa $. We will start by describing a naive heuristic based on breadth-first search (BFS).
	\begin{spec}
		\label{spec:bfs}
		$$ \Phi((A, \mu)) := \Big \{ \arg \min \big \{ \mu(a) : a \in A \setminus \{ g \} \big \} \Big \} $$
	\end{spec}
	
	Using Spec. \ref{spec:bfs}, Alg. \ref{alg:rand} does a BFS over all possible multi-sets. Corollaries \ref{cor:iaps}, \ref{lem:subset} and Lemma \ref{lem:lin0} show that Alg. \ref{alg:rand} terminates for any solvable instance of IAP after $ n |A| $ while-loops, where $ n = \min \{ |l| : l \in \{ S_{0} \rightarrow g \}_{\A} \} $. In each while loop, we solve an ILP with $ O(|\Aa|^2) $ variables and $ O(|\Aa||R|) $ constraints. If $ \{ S_{0} \rightarrow g \}_{\A} = \emptyset $, the algorithm never terminates. Th. \ref{th:nonlin} proves the correctness of Alg. \ref{alg:rand}.

	\section{$ \Phi $-Heuristics for IAP as Search}
	We assume in Alg. \ref{alg:rand} initially that $ \mu(a) = 1 $ for all $ a \in A $, which gives all occurrences of the action $ a $ the same location w.r.t. all other actions $ b \in A $. After that $ \mu(a) $ is increased for all $ a $'s one by one, until an $ l \in \{ S_{0} \rightarrow g \}_{\A} $ is found. However, we only need to increase $ \mu(a) $ for violating actions $ a \in A $ because occurrences of non-violating actions do not have to be ordered w.r.t. each other, i.e., we do not need Boolean repetitions of them. We can do that by:
	$$ \Phi((A, \mu)) = \left \lbrace 
	\arg \min  \left \lbrace  \begin{split} \mu(a) : & \, a \in A \setminus \{ g \}, \\ & \exists b \in A \stt a \sim b \end{split} \right \rbrace
	\right \rbrace $$
	\begin{definition}
		For IAP $ \{S_{0} \rightarrow g\}_{\A} $, if for all $ a, b \in A \setminus \{ g \} $, $ a \nsim b $ holds, we call it an nIAP problem.
	\end{definition}
	\begin{theorem}
		\label{th:novio}
		nIAP is NP-complete.
		
	\end{theorem}
	\begin{proof}
		nIAP can be reduced polynomially to the ILP feasibility problem $ [S_{0} \rightarrow g]_{(A, 1)} $, which is NP-complete. Since no actions violate each other, none of the incomparable occurrences of $ a $ to $ b $ could invalidate the preconditions of $ b $ in any linearization if they occur before or after it. Thus, all occurrences of $ a $ are unordered copies (linear repetitions), and $ \mu(a) = 1 $ suffices, because only one ordered copy (Boolean repetition) of $ a $ is needed, for all $ a \in A $. Thus:
		$$ \{S_{0} \rightarrow g\}_{\A} = [S_{0} \rightarrow g]_{(A,1)} $$
		On the other hand, the ILP-feasibility problem can be reduced polynomially to nIAP. In particular, for ILP:
		$$ \mathbf{M} \mathbf{x} \leq \mathbf{b}, \mathbf{x} \in \N^n, \mathbf{b} \in \Z^m, \mathbf{M} \in \Z^{m \times n} $$
		We define a register $ y_{j} $ for each constraint, an action $ x_{i} $ for each variable $ \mathbf{x}[i] $ that has no preconditions but effects $ \sigma(x_{i},y_{j}) = \mathbf{M}[j, i] $. Additionally, we define the goal $ g $ s.t. $ \upprec(g, y_{j}) = \mathbf{b}[j] $, for $ i \in \{1, ..., n\}, j \in \{ 1, ..., m \} $. If a solution $ l $ of this nIAP is found, then the ILP is feasible by the solution $ \mathbf{x}[i] := \#_{x_{i}}(l) $. \\
	\end{proof}
	We conclude that by ignoring violations, we get a decidable ILP problem $ [S_{0} \rightarrow g]_{(A, 1)} $, which can be used as a heuristic. This will allow us to reduce the number of violations that should be considered, i.e., reduce the number of actions of which additional ordered copies are needed. We will clarify how this heuristic is admissible after defining it formally.
	\begin{definition}
		For an IAP $ \{ S_{0} \rightarrow g \}_{\A} $, a \textit{relaxed MvPOP solution} $ \Pa $ over $ \A $ of $ \{ S_{0} \rightarrow g \}_{\A} $, is an MvPOP over $ \Aa $ with max. $ g $ s.t. $$ \Loprec \leq \Ss_{0} + \Pa \Sigma \leq \Upprec $$
	\end{definition}
	\textit{In Example 5, $ \Pa_{5} $ (compare Fig. \ref{fig:e5}) was a relaxed solution over $ (A_{5}, 1) $ of $ \E_{5} $ but was not a valid one. Specifically, we had $ \Loprec_{5} \leq \Ss_{0} + \Pa_{5} \Sigma_{5} \leq \Upprec_{5} $ but $ \Ss_{0} + (\Pa_{5} + \I_{\leq\F}^{\Pa_{5}}) \Sigma_{5} \not \leq \Upprec_{5} $.}
	\begin{definition}
		For an IAP $ \{ S_{0} \rightarrow g \}_{\A} $, an \textit{integer addition planning relaxed problem} $ ( S_{0} \rightarrow g )_{\Aa} $ is defined as the set of linearizations of all relaxed MvPOPs solutions over $ \Aa $ of $ \{ S_{0} \rightarrow g \}_{\A} $.
	\end{definition}
	\textit{In Example 5, $ \lin(\Pa_{5}) = \{ l^1_{5}, l^2_{5}, l^3_{5} \} $, which contains the right solution $ l_{5}^3 $, because $ \Pa_{5} $ is a relaxed MvPOP solution of $ \E_{5} $.}
	
	In general, in any relaxed MvPOP solution, we guarantee that the preconditions of at least one occurrence of each action $ a \in \Aa $ are satisfied, meaning that any valid solution is also a relaxed one, and that it is never harder to solve the relaxed version. Thus, for any IAP $ \{ S_{0} \rightarrow g \}_{\A} $, and any multi-set $ \Aa = (A, \mu) $, s.t. $ \mu(a) \geq 1 $ for all $ a \in A $:
	$$ \{S_{0} \rightarrow g\}_{\A} \subseteq (S_{0} \rightarrow g)_{\Aa} $$
	In words, a heuristic using this relaxation is admissible. Moreover, such a heuristic is also useful in another way because although many IAPs have an infinite number of solutions, when minimality is required, only specific solutions with similar used actions are considered. Since:
	$$ \min \{ |l| : l \in \{ S_{0} \rightarrow g \}_{\A} \} \geq \min \{ |l| : l \in ( S_{0} \rightarrow g )_{\Aa} \} $$
	Then, we can suggest that if an action $ a $ is used $ n $ times in the minimal relaxed solution, then, it will be used at least $ n $ times in the minimal valid solution.
	\begin{spec}
		\label{spec:opt}
		Let $ \Pa $ be a minimal relaxed MvPOP solution over $ \Aa $ with max. $ g $:
		$$ \Phi(\Aa) := \{ a \in \Aa : \existst b \in \Aa \stt a \sim_{\Pa} b \} $$
		Where $ a \sim_{\Pa} b $ iff. $ (a, b) $ causes a threat in $ \Pa $, i.e., there exists $ x \in R $ and $ \circ \in \{\leq, \geq \} $ s.t. $ \Pa[g, a] > \I_{\circ x}^{\Pa_{m}}[b, a] > 0 $, where $ \Ss[b, x] \! \not \! \circ \; \Pi^\circ[b, x] $, for $ \Ss :=  \Ss_{0} + (\Pa + \I_{\circ x}^{\Pa_{m}}) \Sigma $.
	\end{spec}
	\textit{In Example 5, $ \Pa_{5} $ is a relaxed minimal solution of $ \E_{5} $. We had $ 0 < \I^{\Pa_{5}}_{\leq\F}[e(1, 3), u] = 2 < \Pa_{5}[g, u] = 3 $, and $ \Ss[e(1, 3), \F] = 3 \not \leq 1 $ (compare Fig. \ref{fig:e5}). Therefore, following Spec. \ref{spec:opt}, we need to increment $ \mu(u) $ to get an additional ordered copy of $ u $. This leads us directly to the correct multi-set with two ordered copies of $ u $ without needing to test whether additional ordered copies of the other actions are required.}
	
	Notice that, with Spec. \ref{spec:opt}, if $ \Pa_{m} $ is a relaxed solution of the IAP $ \{ S_{0} \rightarrow g \}_{\A} $, then, $ \lin(\Pa_{m}) \subseteq (S_{0} \rightarrow g)_{\Aa} $. If the IAP is infeasible, then $ \{ S_{0} \rightarrow g \}_{\A} \subseteq (S_{0} \rightarrow g)_{\Aa} = \emptyset $. In this case, we get $ \Phi(\Aa) = \emptyset $, and the search is aborted anyway.
	\section{NP-complete Fragment of IAP}
	In this section, we use the heuristic defined before to prove the existence of an NP-complete fragment of IAP.
	\begin{definition}
		In an action set $ A $, we call an action list $ [a_{0}, a_{2}, ..., a_{n}] \in A^* $ a $ \sim $-\textit{cycle} of length $ n $, iff. $ a_{i-1} \sim a_{i} $ for all $ i $ in $ \{1, ..., n\} $, and $ a_{n} = a_{0} $.
	\end{definition}
	
	\begin{definition}
		For IAP $ \{ S_{0} \rightarrow g \}_{\A} $, if $ k $ is the maximal length of of a $ \sim $-cycle in $ A $, we call $ \{ S_{0} \rightarrow g \}_{\A} $ a $ k $-IAP problem.
	\end{definition}
	
	\textit{In the offline elevator problem, $ u \sim u $ forms the longest violation cycle: $ [u, u] $, which is of length $ 1 $, i.e. the elevator problem is $ 1 $-IAP.}
	
	\begin{theorem}
		\label{th:iap0}
		For any $ 0 $-IAP $ \{ S_{0} \rightarrow g \}_{\A} $: $$ (S_{0} \rightarrow g)_{(A, 1)} \neq \emptyset \impliest \{ S_{0} \rightarrow g \}_{\A} \neq \emptyset $$
	\end{theorem}
	\begin{proof}
		Since $ (S_{0} \rightarrow g)_{(A, 1)} \neq \emptyset $, then, there exists a minimal MvPOP $ \Pa $, such that $ \lin(\Pa) \subseteq (S_{0} \rightarrow g)_{(A, 1)} $. We will prove that $ \lin(\Pa) \cap \{ S_{0} \rightarrow g \}_{\A} \neq \emptyset $. We know that $$ |\lin(\Pa)| = \prod_{a \in A} \sum_{\substack{b \in A \\ b \neq a}} (\I^\Pa[b, a] + 1) $$
		because only incomparable occurrences have some freedom to be located differently, as discussed in Def. \ref{def:inc0}.	For a pair of actions $ c \sim_{\Pa} d $, we delete all linearizations where any part of the violating incomparable action copies of $ c $ to $ d $ occur before $ d $ (similar to increasing $ \mu(c) $ in Algorithm \ref{alg:rand}, which allows for the distinction of $ c $'s occurrences before and after $ d $). In other words, we \textit{promote} the violating incomparable occurrences of $ c $ over $ d $. We iterate this over all pairs $ c \sim_{\Pa} d $ and get $ L \subseteq \lin(\Pa) $:
		$$
		|L| = \prod_{a \in A} \sum_{\substack{b \in A, b \neq a \\ a \not \sim_{\Pa} b}} (\I^{\Pa}[b, a] + 1) $$
		Notice that $ L \subseteq \{ S_{0} \rightarrow g \}_{\A} $, because we dealt with all violations in a relaxed MvPOP solution that ignores violations by promoting all violating incomparable occurrences, which makes all linearizations left valid if there are any. \\ Finally, notice that $ \sim_{\Pa} $ is acyclic, because $ a \sim_{\Pa} b $ implies $ a \sim b $, and there are no $ \sim $-cycles. Therefore, there exists $ b \in A $ s.t. $ a \not \sim_{\Pa} b $ for all $ a \in A $. Thus, $ |L| > 0 $.
	\end{proof}
	
	\textit{In Example 5, consider $ \lin(\Pa_{5}) = \{ l^1_{5}, l^2_{5}, l^3_{5} \} $: If we delete all linearizations where any $ u $ incomparable occurrences to $ e(1, 3) $ occur before $ e(1, 3) $(i.e., promote the two incomparable occurrences of $ u $ over $ e(1, 3) $), we delete $ l^1_{5}, l^2_{5} $ and we are left with $ l^3_{5} $, which is a valid solution of $ \E_{5} $. Remember that $ [u, u] $ is a violation cycle in the offline elevator problem, which means it is not a $ 0 $-IAP instance. However, because in the given relaxed solution $ \Pa_{5} $ of $ \E_{5} $, only $ u \viogeq_{\F} e(1, 3) $ causes a threat, i.e., $ u \sim_{\Pa_{5}} e(1, 3) $ but $ u \nsim_{\Pa_{5}} u $ (Fig. \ref{fig:e5}); thus, the cycle $ u \sim u $ is not relevant and the algorithm described in Th. \ref{th:iap0} works in that case as well. This is an example of how the heuristic described in Spec. \ref{spec:opt} decreases the number of violations needed to be dealt with.}
	\begin{corollary}
		$ 0 $-IAP is NP-complete.
	\end{corollary}
	\begin{proof}
		Notice that any MvPOP solution of a $ 0 $-IAP problem is of polynomial size (size of $ |\Aa| $ w.r.t. $ |A| $) because we start with $ |(A, 1)|=|A| $ and increment $ \mu(a) $(described in Th. \ref{th:iap0} as promotion or linearization elimination) at most $ O(|A|^2) $ times for all $ a \in A $ because the maximal number of edges in an acyclic graph with $ n $ nodes is in $ O(n^2) $. \\
		From Theorem \ref{th:novio}, $ 0 $-IAP is NP-hard.
	\end{proof}
	\begin{corollary}
		$ 1 $-IAP is NP-complete.
	\end{corollary}
	\begin{proof}
		Notice that the proof of Theorem \ref{th:iap0} does not mention the violating incomparable occurrences of any action $ a $ to itself because this requires a $ \sim $-cycle of length $ 1 $. Let $ \Pa_{0} $ be a relaxed solution of a $ 1 $-IAP $ \{S_{0} \rightarrow g\}_{\A} $. Let $ a \sim_{\Pa_{0}}^* b $ iff. $ a \sim_{\Pa_{0}} b $ and $ a \neq b $ in $ \Pa_{0} $; thus, there exists a maximal $ b $ w.r.t. $ \sim_{\Pa_{0}}^* $. Then, $ \Pa_{0}[g, b] $ is an upper bound for the number of needed ordered copies of $ b $ because $ b $ only violates itself but no other actions in the relaxed solution $ \Pa_{0} $. Now, let $ \Aa_{1} = (A, \mu_{1}) $ s.t. $ \mu_{1}(b) = \Pa_{0}[g, b] $, i.e., $ \Aa_{1} $ has as many ordered copies of $ b $ as there are unordered ones in $ \Pa_{0} $. Then, in a relaxed solution $ \Pa_{1} $ over $ \Aa_{1} $, $ \I^{\Pa_{1}}[b, b] = 0 $ because each ordered copy of $ b $ in $ \Aa $ occurs only once in $ \Pa_{1} $. Therefore, the number of $ \sim_{\Pa_{1}} $-cycles is strictly less than the number of $ \sim_{\Pa_{0}} $-cycles. Iterating this over all actions will eventually lead to either finding a solution of $ \{S_{0} \rightarrow g\}_{\A} $ or conclude that no such solution exists because $ \mu(b) $ does not have to be incremented any further. This produces a polynomial-size MvPOP as well because the number of unordered copies needed in a relaxed solution (e.g., $ \Pa_{0}[g, b] $ above) is only dependent on the preconditions but not on the size of $ A $ or $ R $.
	\end{proof}
	\begin{example}
		We can find out that $ \E_{6} = (0,1,\{(0,2)\}) $ is unsolvable by first solving the relaxed problem and get $ \Pa $ s.t. $ \lin(\Pa) = [e(0,2), u, u, l(0, 2), g] $(i.e., $ \Pa[g, u] = 2 $). Then, we conclude that $ u \sim_{\Pa} u $, because an elevator can never move up twice successively in a building with two floors. However, no other action violates $ u $. Thus, $ \Pa[g, u] = 2 $ is an upper bound for the number of needed ordered copies of $ u $. We do not need to search further for multi-sets with more ordered copies of $ u $.
	\end{example}
	\section{$ f $-Optimization for IAP as Search}
	Let us finally discuss the function $ f(\Pa) $, which represents the minimization objective of the ILP system in Alg. \ref{alg:rand}.
	\subsection{Minimal Cost}
	If we have a function $ c: A \rightarrow \mathbb{R} $ that defines the cost of each action, we can use it to minimize the costs or the number of used actions. In the offline elevator example, we define $ c(u) = c(d) = 1 $, and $ c(a) = 0 $ for all other actions $ a $, to minimize the energy used by the elevator.
	\begin{spec}
		For an MvPOP $ \Pa $ over $ \Aa $ with max. $ g $:
		$$ f(\Pa) := \sum_{a \in \Aa} c(a) \Pa[g, a] $$
	\end{spec}
	If $ c(a) = 1 $ for all $ a \in A $, we get a minimal length plan, which is needed for the heuristics described before.
	\subsection{Defaults}
	In knowledge representable planning, a \textit{Boolean default} $ d $ represents a soft precondition, i.e., a precondition that should be fulfilled as long as fulfilling it does not invalidate any (hard) precondition. Again, from a (linear) numeric perspective, if we say that $ e(d) = 1 $ if $ d $ is $ \top $ and $ 0 $ otherwise, then a default can be modeled as an ILP problem ``maximize $ e(d) \in \N $ s.t. $ e(d) \leq 1 $ and all hard preconditions hold". This makes it a \textit{linear default}. There are examples of linear defaults, like telling the elevator to move up as many floors as possible before moving down again and vice versa, which is a common preference for elevators. Notice that linear defaults can model Boolean defaults; therefore, we need to discuss linear defaults only.
	\begin{definition}
		An \textit{integer addition action description with defaults} is a tuple $ (\A, \delta, \rho) $, where $ \A $ is an integer addition description (IAD), $ \delta : A \times R \rightarrow \{ +1, -1\} $ is the default function, and $ \rho : A \times R \rightarrow \N $ is the relevance function.
	\end{definition}
	By $ \delta(b, x) = +1 $, we mean that $ b $ needs the value of $ x $ to be as big as possible before applying it. This must be bounded by some other (hard) preconditions, or else we will get infinite plans. Similarly, $ \delta(b, x) = -1 $ means that $ b $ needs the value of $ x $ to be as small as possible. The value $ \rho(b, x) $ defines the relevance of the default $ \delta(b, x) $. This allows for preferences among defaults because it allows for comparison between them. An irrelevant default is represented by $ \rho(b, x) = 0 $. The more important the default, the higher its relevance value. \\
	\begin{spec}
		\label{spec:def}
		For an MvPOP $ \Pa $ over $ \Aa $ with max. $ g $:
		\begin{equation*}
			\begin{split}
				f(\Pa) := & \sum_{a \in \Aa} \Pa[g, a] \\
				& - 2\sum_{x \in R} \sum_{a, b \in \Aa} \rho(b, x) \delta(b, x) \Pa[b, a]  \sigma(a, x)
			\end{split}
		\end{equation*}	
	\end{spec}
	The objective is to minimize $ f(\Pa $). Thus, using Spec. \ref{spec:def}, we guarantee that additions on register $ x $ before an action $ b $ with default $ \delta(b, x) = +1 $ are maximized by maximizing the number of occurrences of all actions $ a $ before $ b $, i.e., $ \Pa[b, a] $, if $ \sigma(a, x) > 0 $, and minimize $ \Pa[b, a] $ if $ \sigma(a, x) < 0 $, both weighted by the incremented / decremented value $ |\sigma(a, x)| $. The opposite happens in the other case, where $ \delta(b, x) = -1 $. Additionally, we favor defaults with higher relevance by multiplying with $ \rho(b, x) $. The first sum guarantees minimality w.r.t. plan length; thus, we multiply the second sum by 2 to cancel the minimization effect.
	
	\section{Conclusion}
	Let us compare the cases where linear and Boolean repetitions are needed (Table \ref{tab:comp}).
	\begin{table}
		
		\centering
		\footnotesize
		\begin{tabular}{|c|c|c|c|}
			\hline
			& Repetitions & Longest $ \sim $-cycle & Complexity \\
			\hline
			STRIPS      & B           & $ * $                      & PSPACE-C     \\ \hline
			nIAP   & L           & $ * $                      & NP-C           \\ \hline
			$ 1 $-IAP       & L, B        & $ 1 $                      & NP-C     \\ \hline
			IAP         & L, B        & $ * $                      & Undecid. \\ \hline  
		\end{tabular}
		\caption{A comparison between the cases where linear (L) and Boolean (B) repetitions are needed.}
		\label{tab:comp}
	\end{table}
	We can see that the undecidability of IAP comes from the need for both linear and Boolean repetitions when there are cycles of violation between different actions at the same time. Additionally, calculating the number of needed linear repetitions is easier than calculating the number of needed Boolean repetitions, especially when cycles of violations are allowed. Finally, even for IAP instances with a relaxed solution that contains no violation cycles between different actions, the problem is NP-complete.
	
	It is worth mentioning that the described algorithms can produce simultaneous plans as well because an MvPOP can be easily translated into a partial order. In the offline elevator problem with two elevators, the algorithm divides the work on two elevators if that reduces the costs, for example. Finally, increasing the number of soft preconditions (or defaults) as well as requiring minimality, do not affect the complexity of the problem.
	\section{Future Work}
	For an MvPOP to represent a cycle, we need ordered copies for each cycle resulting in a redundant representation. To clarify this, we analyze the \textit{drinking water} example. One can only drink ($ d $) when the glass is filled. One can only fill the glass ($ f $) when it is empty. Both $ f $ and $ d $ violate themselves but need each other. Initially, the glass is empty. The goal is to drink $ i $ times. Notice that there are two $ \sim $-cycles of length 1. For this problem, we need an MvPOP over a multi-set of cardinality $ 2i + 1 $ actions because the minimal solution is $ [f, d, f, d, .., g] $. A naive solution to this problem is to define concatenation for actions and concatenate when needed. For example, solving $ \{ S_{0} \rightarrow d \}_{\A} $ results in $ l = [f, d] $, which can be used as one action later $ fd $. For the drinking example, the IAP $ \{S_{0} \rightarrow g \}_{\A} $ is solvable by an MvPOP over the multi-set $ \{ fd, g \} $, which has two actions. We are developing this idea further to study the complexity of $ k $-IAP for fixed $ k $. We know, though, that the introduction of action concatenation is similar to the introduction of subroutines which is known to make problems undecidable. We believe this research can make the decidability edge between 1-IAP and IAP sharper, leading to new insights in numeric planning.
	
	\section{Acknowledgment}
	The authors gratefully acknowledge the funding by the BMBF in Germany for the project AIStudyBuddy (No. 16DHBKI016), and the EU ICT-48 2020 project TAILOR (No. 952215).
	
	\bibliographystyle{kr}
	\bibliography{kr23-instructions.bib}
\end{document}